\documentclass[11pt]{article}
\usepackage[preprint]{acl}
\usepackage{times}
\usepackage{latexsym}
\usepackage[T1]{fontenc}
\usepackage[utf8]{inputenc}
\usepackage{microtype}
\usepackage{booktabs}
\usepackage{amsmath}
\usepackage{multirow}
\usepackage{graphicx}

\newcommand{\email}[1]{{\fontsize{11pt}{11pt}\selectfont\texttt{#1}}}

\title{Context-Grounding Gains Are Mediated by Pre-existing Machinery:\\ Auditing GRPO, SFT, and DPO}

\author{Prakhar Gupta \\
  University of Michigan \\
  \email{prakharg@umich.edu} \\ \And
  Vaibhav Gupta \\
  University of Waterloo \\
  \email{v223gupta@uwaterloo.ca} \\}

\begin{document}
\maketitle

\begin{abstract}
Language models can ignore prompt evidence when it conflicts with memorized
knowledge. Post-training can make models follow such evidence more reliably,
but it is unclear whether these gains require new machinery or strengthen
machinery already present. We compare nine post-training arms spanning GRPO,
SFT, and DPO from one starting checkpoint, with key comparisons extended across scales and families. We estimate a grounding direction from that
checkpoint before training. Across five tested GRPO variants, grounding gains
are small. For the two variants replicated across seeds, equivalence tests
bound their effects below the conflict-SFT gain even as the rewarded metric
improves. Conflict-SFT improves grounding moderately, while DPO drives
grounding near ceiling on its matched distribution. Conflict-SFT and DPO
largely use the same causal attention-head set as the starting model.
Subtracting the starting-model direction suppresses both gains, while adding it
to the starting model recovers 35\% of DPO's gain at a dose passing all stated
side-effect checks. After a supervised warm start makes the context answer
appear in more rollouts, the same GRPO recipe adds essentially no further
grounding gain. In our setting, grounding gains largely depend on machinery
already present in the starting model.
\end{abstract}

\section{Introduction}
\label{sec:intro}

Language models can ignore evidence in the prompt when it conflicts with
memorized knowledge \citep{longpre2021entity,ming2025faitheval}. We study this
failure as \textbf{context grounding under knowledge conflict}, where prompt
evidence contradicts the model's parametric memory. Does the model follow the
supplied evidence or its memorized answer? Post-training can improve context
grounding \citep{li2022kaft,bi2024contextdpo}, but it is less clear what changes
inside the model. When training improves grounding, does it build new internal
machinery or strengthen machinery already present in the starting model?

Prior work shows that the latter can occur. Fine-tuning can preserve and
strengthen existing mechanisms \citep{prakash2024finetuning}, and the choice
between context and memory can be controlled by a low-dimensional direction
that also works in models without task-specific fine-tuning
\citep{minder2025controllable}. DPO can likewise change behaviour while leaving
pre-existing capabilities recoverable \citep{lee2024mechanistic}. This leaves
a more specific question. When several post-training recipes start from the
same model, do their gains depend on machinery that can already be identified
before training?

At Qwen2.5-1.5B, we train nine arms, including five GRPO variants, three SFT
variants, and DPO. We extend key comparisons to larger Qwen models and to Llama and Phi. These are complete training recipes, so objective, data, and
budget can vary together. Our comparisons therefore concern the tested recipes
rather than the optimization objective alone. Before training, we estimate a
grounding direction from the starting model. We combine this with causal
attention-head analysis to test what later training changes.

\begin{figure*}[t]
\centering
\includegraphics[width=\textwidth]{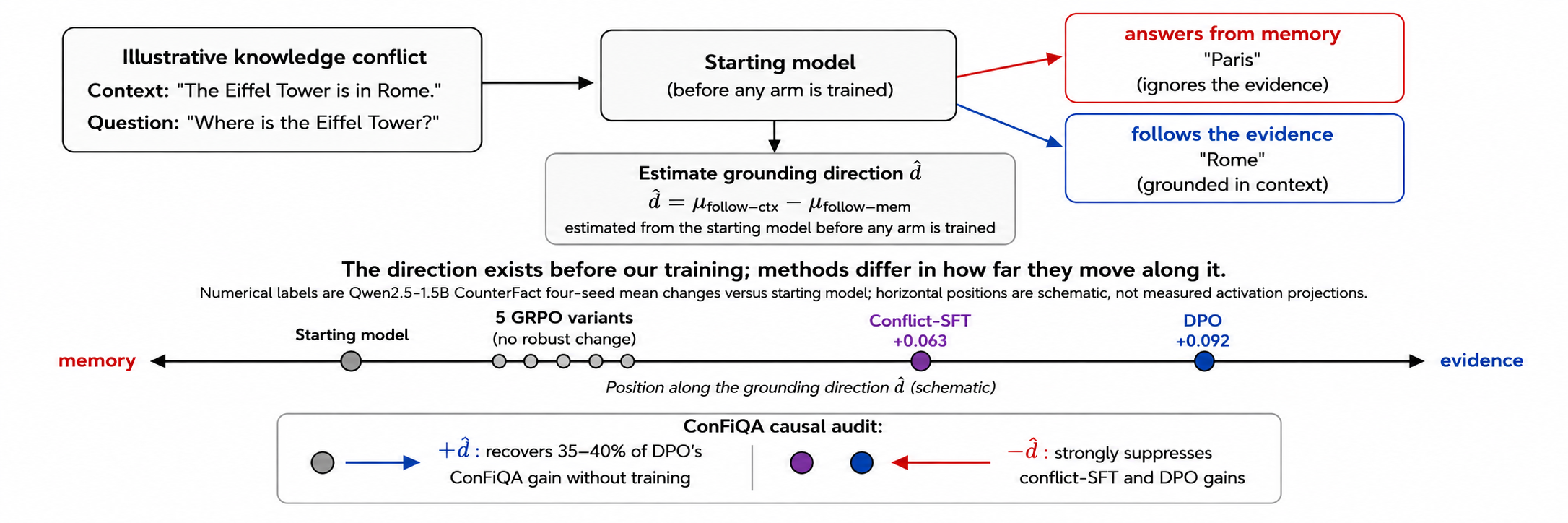}
\caption{\textbf{Post-training methods differ in grounding gains, while a
grounding direction can already be identified in the starting model.}
The direction is estimated from the starting model alone before any arm is
trained. Adding it increases grounding in the starting model, while subtracting
it suppresses conflict-SFT and DPO gains. Prompt formats are in
Appendix~\ref{app:prompts}, the steering recovery with its dose and side-effect
checks is in \S\ref{sec:steer}, and the other main effects are in
\S\ref{sec:behav}--\S\ref{sec:mech} and Appendix
Figure~\ref{fig:quant}.}
\label{fig:hero}
\end{figure*}

\paragraph{Findings.}
\textbf{(1) The training recipes differ sharply in grounding gains}
(\S\ref{sec:behav}). Across five tested GRPO variants, grounding gains are
small. For the two seed-replicated variants, equivalence tests bound their
effects below the conflict-SFT gain, even though the rewarded metric improves.
Conflict-SFT improves grounding moderately ($+.044$ over its matched control),
while DPO drives grounding near ceiling on its matched distribution
($+.36$--$+.60$ on ConFiQA across five models in three families).

\textbf{(2) The gains that do appear largely depend on machinery already
present in the starting model} (\S\ref{sec:mech}). Across the audited arms,
independently discovered causal head sets largely recover the starting model's
top heads (7--8/8 at 1.5B), while a matched recall task shares 0/8 and
cross-task ablations are strongly asymmetric. The grounding direction estimated
from the starting model remains closely aligned with directions in the trained
models. Subtracting it suppresses the conflict-SFT and DPO gains, while adding
it to the starting model recovers 35\% of DPO's gain at a dose that passes all stated side-effect checks.

\textbf{(3) The grounding direction remains stable as the behaviour improves}
(\S\ref{sec:traj}, \S\ref{sec:coverage}). During DPO training, grounding reaches
about 90\% of its final level by step 160 of 800 while the grounding direction
remains closely aligned with its starting orientation (cosine $\ge .968$). We
also test a simple explanation for the GRPO result. A supervised warm start makes the context answer appear in more rollouts, but
the same GRPO recipe then adds essentially no further grounding gain
($+.001$, $p{=}.91$).

Together, these results support a scoped mechanism-reuse account of
context-grounding post-training. We do not claim novelty for mechanism reuse or
for the context-versus-memory direction. Our contribution is to compare
multiple post-training recipe families on the same grounding behaviour, estimate
the grounding direction in the common starting model before training, and then
test reuse causally through independent head localization and interventions
that increase grounding in the starting model and suppress later conflict-SFT
and DPO gains. Throughout, ``pre-existing'' means present in the
instruction-tuned starting checkpoint, not necessarily in the raw pretrained
model.

\section{Related Work}
\label{sec:related}

\paragraph{Context grounding and post-training.}
Language models often rely on parametric knowledge even when the prompt provides
conflicting evidence \citep{longpre2021entity,ming2025faitheval}. Prior work has
localized mechanisms involved in this choice
\citep{ortu2024competition,minder2025controllable,jin2024cutting} and improved
context use through prompting, decoding, steering, and fine-tuning
\citep{zhou2023context,shi2024trusting,contextfocus2026,li2022kaft,
bi2024contextdpo}. Recent work also tracks how context sensitivity changes
through SFT, DPO, and RLVR \citep{context_usage_2026}. More broadly, recent work
asks how RL gains relate to capabilities or representations available before RL
\citep{yue2025does,liu2025prorl,invisibleleash2025,gupta2026worldmodels}. RL can improve
grounding when the reward directly targets evidence use.
\citet{chen2026longrlvr} add a verifiable context reward to outcome-based RLVR,
while \citet{tamo2026evidencerl} train GRPO with an explicit evidence-grounding
reward. Our GRPO result is therefore specific to the tested GRPO recipes and
starting policy, not a claim that RL cannot improve grounding. Together, this
work shows that context use can be localized, manipulated, and trained. A
remaining question is whether different post-training recipes starting from the
same checkpoint obtain their grounding gains through shared machinery that can
already be identified before training.

\paragraph{Mechanism reuse under post-training.}
Several studies show that post-training can reuse structure already present in
the starting model. \citet{prakash2024finetuning} find that entity-tracking
improvements largely preserve the original circuit, while
\citet{lee2024mechanistic} find that DPO reduces toxicity by bypassing rather
than removing pre-existing capabilities. A context-versus-memory direction
identified after task fine-tuning also controls non-fine-tuned models
\citep{minder2025controllable}. Base-model representations can also be
repurposed by reasoning fine-tuning \citep{ward2025reasoning} or recruited as a
persona subspace identified before fine-tuning \citep{persona_subspace_2026}.
Recent studies report both preservation and substantial representational change
across post-training settings
\citep{bigoulaeva2026patches,shi2026features,gupta2026biasdirections}. Our study
combines these questions for context grounding. We compare GRPO, SFT, and DPO from a common starting checkpoint and estimate the
grounding direction before any of them are trained. We test reuse causally
through independent head localization and by using the same starting-model
direction to both increase grounding in the starting model and suppress
conflict-SFT and DPO gains after training.

\paragraph{Activation steering.}
Activation steering provides a direct way to test whether a representation has
causal control over behaviour \citep{rimsky2024caa,turner2023actadd,
arditi2024refusal,zou2023repe}. Steering can also be unreliable or non-specific
\citep{tan2024analysing,pres2024towards,dim_specificity_2026}. We therefore
compare against matched-norm random directions, report per-input helped and hurt
fractions, and check KL and general capabilities. We use steering as a causal
test of the pre-existing grounding direction, not as a replacement for
post-training.

\section{Setup}
\label{sec:setup}

\paragraph{Models and training arms.}
Base models: Qwen2.5-1.5B/3B/7B-Instruct, Llama-3.2-3B-Instruct, Phi-3.5-mini-Instruct.
Throughout, ``base'' means this instruction-tuned starting checkpoint.
Pre-existence claims are relative to it, not to raw pretraining (Limitations).
The arms compare complete training recipes as practitioners deploy them, so objective, data, and budget vary together by design.
At 1.5B we train nine arms (Table~\ref{tab:zoo}).
Five are GRPO variants \citep{shao2024deepseekmath}: \textbf{A} (answer token-F1 reward), \textbf{A$'$} (same reward, evidence-emphasizing prompt), \textbf{B} (adds a citation set-F1 reward), \textbf{C} (adds a contrastive context-utility reward, $\log p(\text{ans}\mid q,\text{ctx})-\log p(\text{ans}\mid q)$), and \textbf{D} (the reward is the in-context counterfactual answer).
Three are SFT arms: \textbf{E0} (no-conflict control), \textbf{E2} (KAFT-style mixture, \citealp{li2022kaft}), and \textbf{E3} (78\% conflict, 22\% standard).
The ninth is \textbf{DPO} \citep{rafailov2023direct} on ConFiQA preference pairs.
GRPO uses 200 steps, lr $3\cdot10^{-6}$, KL $\beta{=}.02$, and group size 8 (TRL).
SFT uses 2 epochs over the same 1{,}500-example pool GRPO samples from.
DPO uses $\beta{=}.5$, 3 epochs, and 4{,}500 pairs.
A, A$'$, E2, E0, E3, and DPO are replicated over 4 seeds at 1.5B.
A, E0, E3, and DPO are additionally trained at 3B (4 seeds each) and at Llama-3.2-3B (single seed), and DPO at 7B and Phi-3.5-mini (single seed).
GRPO arms train on clean QA (HotpotQA, \citealp{yang2018hotpotqa}) except D, which trains on conflict data with the context answer as its reward.
For all variants except D, the reward does not directly target grounding.
This lets us test whether optimizing related task rewards also improves
grounding.

\paragraph{Evaluation.}
The CounterFact conflict protocol \citep{meng2022locating} uses a two-pass design.
Pass 1 (closed book) finds items the model answers correctly from memory.
This \textbf{frozen known set} ($n{=}1089$ at 1.5B, per-model at other scales) is computed once from the base model and reused for every arm, so all comparisons are paired on identical items.
Pass 2 presents a counterfactual context, and we record whether the generation contains the context answer (\emph{follow-ctx}) or the memorized answer (\emph{follow-mem}).
The headline metric is the \textbf{update rate} $=$ follow-ctx$/$(follow-ctx$+$follow-mem).
ConFiQA \citep{bi2024contextdpo} and FaithEval \citep{ming2025faitheval} provide replications that differ in distribution and format (free-form and MCQA).
Their known sets are computed per model, so paired tests there use items known in both models, with $n$ reported per test.
Statistics: exact two-sided McNemar on paired items.
Seed-level CIs use $t(.975,n{-}1)$ with $n{=}4$.
Equivalence is tested via two one-sided tests (TOST) using $\pm.044$, the
observed conflict-SFT gain over its matched control, as the equivalence margin.
Single-seed $p$-values in tables are descriptive. Seed-level CIs are reported
for the replicated comparisons, and the single-seed extensions are identified
above.

\paragraph{Metric integrity under intervention.}
We flag one measurement issue. Under an activation intervention, both the known-set gate and the ``decisive''
subset can change with the intervention, so ratio metrics conditioned on them
can be distorted.
In our audit, this bias changed one effect estimate by a factor of four.
For every steered or projected run we therefore report the \textbf{exclusive follow-ctx rate over the frozen no-intervention known set} (denominator fixed before the intervention exists), with the non-decisive fraction reported separately.
Appendix~\ref{app:metric} details the failure mode.

\paragraph{Direction estimation.}
For the main 1.5B audit, the direction is a difference-in-means (DiM): mean
last-position residual on follow-ctx items minus follow-mem items, computed on
the base model at layer 21 of 28.
Selection protocol: at 1.5B the layer is the probe-selectivity optimum (\S\ref{sec:mech}), and every intervention at it is validated against matched randoms.
At 3B the locus was located by a full layer sweep against matched randoms, with
an even/odd item split used to check for overfitting to the evaluation items.
Steering adds $\alpha\hat d$ at the last position during generation, removal uses $\alpha<0$, and $\alpha$ is norm-calibrated across families (Llama-3.2-3B mean $\lVert h\rVert{=}26.7$ vs.\ Qwen $103.7$).
All interventions are compared against matched-norm random directions, following the specificity controls of \citet{dim_specificity_2026}.

\section{A Three-Way Behavioural Dissociation}
\label{sec:behav}

Comparing nine training recipes from one starting checkpoint, grounding gains
differ sharply. The tested GRPO recipes yield small gains, conflict-SFT yields
moderate gains, and DPO yields the largest gain on CounterFact and near-ceiling
grounding on its matched ConFiQA distribution
(Tables~\ref{tab:zoo}--\ref{tab:dpo}).

\begin{table}[t]
\centering\small
\begin{tabular}{@{}llrrr@{}}
\toprule
Arm & Objective & Rate & $\Delta$ & $p$ \\
\midrule
base & --- & .558 & --- & --- \\
\midrule
A & GRPO, answer F1 & .569 & $+.011$ & .62 \\
A$'$ & GRPO, evid.\ prompt & .595 & $+.036$ & .007 \\
B & GRPO, $+$citation & .563 & $+.004$ & .63 \\
C & GRPO, $+$ctx-utility & .577 & $+.018$ & .027 \\
D & GRPO on conflict & .566 & $+.008$ & .19 \\
\midrule
E0 & SFT control & .583 & $+.025$ & .09 \\
E2 & SFT, KAFT mix & .602 & $+.044$ & .003 \\
E3 & SFT, conflict & .620 & $+.062$ & $1.2\mathrm{e}{-7}$ \\
DPO & pref.\ pairs & .646 & $+.087$ & $1.0\mathrm{e}{-13}$ \\
\midrule
\multicolumn{5}{@{}l@{}}{4-seed mean $\Delta$ [95\% $t$-CI]:}\\
\multicolumn{5}{@{}l@{}}{\quad A $+.006\;[-.000,+.012]$}\\
\multicolumn{5}{@{}l@{}}{\quad A$'$ $+.017\;[-.008,+.041]$}\\
\multicolumn{5}{@{}l@{}}{\quad E2 $+.047\;[+.041,+.054]$}\\
\multicolumn{5}{@{}l@{}}{\quad DPO $+.092\;[+.084,+.099]$}\\
\bottomrule
\end{tabular}
\caption{\textbf{GRPO grounding gains are small, while SFT and DPO improve
grounding more strongly.} CounterFact update rate on the frozen known set
($n{=}1089$, paired exact McNemar vs.\ base). For GRPO A and A$'$, the
4-seed CIs include zero.}
\label{tab:zoo}
\end{table}

\paragraph{GRPO yields small grounding gains while improving its rewarded
metric.}
No GRPO variant moves grounding robustly (Table~\ref{tab:zoo}).
Holm-corrected within the five-variant family, only A$'$ retains nominal
significance ($p{=}.035$, with C at $p{=}.11$, not seed-replicated), and the two
seeded GRPO variants shrink under replication, with both 4-seed CIs including
zero. For A and A$'$, equivalence tests support effects within $\pm.044$, the
observed conflict-SFT gain over its matched control (A $p{=}.0001$, A$'$
$p{=}.0197$). We therefore claim a bound rather than a zero effect for these
seeded variants. The single-seed B, C, and D point estimates are all below
$+.02$.

The small grounding gains do not reflect a failure of training. The same
answer-F1 runs raise HotpotQA F1 by $+.120$ at 1.5B ($.442{\to}.563$) and
$+.104$ at 3B ($.566{\to}.670$), so training improves the metric it directly
rewards. The same pattern appears at Qwen-3B (4-seed mean
$\Delta{=}-.002$, CI $[-.008,+.004]$) and in a single Llama-3.2-3B run
($.596{\to}.598$, $p{=}.47$, with F1 $+.081$). The largest variant effect
(A$'$, prompt-format) averages $+.017$, is seed-variable, and does not transfer
cross-dataset (ConFiQA $.590$ and FaithEval $.689$, both $\approx$ base). This
is consistent with template-specific elicitation
\citep{liu2025drgrpo,shao2025spurious,yue2025does} rather than a robust
grounding improvement.

The claim is limited to on-policy GRPO from a base policy with low
context-answer rollout coverage on its training distribution. Methods that use synthetic data and tailored grounding rewards
\citep{si2025canoe} fall outside this setting. \citet{cot_grpo_2025} report GRPO beating DPO for CoT
faithfulness at 14B, a different behaviour at a scale where the authors note
small-model instability, while our range is 1.5--7B.

\begin{table}[t]
\centering\small
\begin{tabular}{@{}lccc@{}}
\toprule
Model & base & $+$DPO & $\Delta$ \\
\midrule
Qwen2.5-1.5B & .586 & .964 & $+.378$ \\
Qwen2.5-3B & .362 & .961 & $+.599$ \\
Qwen2.5-7B & .446 & .969 & $+.524$ \\
Llama-3.2-3B & .492 & .980 & $+.488$ \\
Phi-3.5-mini & .586 & .942 & $+.356$ \\
\bottomrule
\end{tabular}
\caption{\textbf{DPO produces near-ceiling grounding on all five models across three families.} ConFiQA-QA update rate (DPO's training distribution), base vs.\ DPO. Paired McNemar $p\le 2.5\mathrm{e}{-25}$ everywhere. Seed $t$-CIs at 1.5B and 3B: $+.376\,[+.370,+.382]$ and $+.596\,[+.591,+.600]$.}
\label{tab:dpo}
\end{table}

\paragraph{SFT improves grounding moderately, DPO nearly to ceiling on matched data.}
Conflict-SFT (E3) adds $+.062$ over base (4-seed mean $+.063$) and $+.044\,[+.034,+.055]$ over its matched no-conflict control, paired by seed.
The KAFT mixture (E2) is similar against base
($+.047\,[+.041,+.054]$) and transfers best among the SFT arms to ConFiQA
($.904$, 47 of 52 decisive items).
DPO produces the largest gains in every CounterFact and ConFiQA comparison
(Table~\ref{tab:dpo}), consistent with the original Context-DPO result
\citep{bi2024contextdpo}. The DPO gain also appears across scales, families,
ConFiQA splits (3B: QA $+.599$, MR $+.471$, MC $+.535$), and on FaithEval
(full table in Appendix~\ref{app:faitheval}).
HotpotQA F1 is essentially unchanged at 1.5B ($.444$ vs.\ $.442$), decreases
from $.566$ to $.536$ at 3B, and has its largest reported drop on Llama
($-.095$).

\section{Evidence for Reuse of Pre-existing Causal Machinery}
\label{sec:mech}

These behavioural differences leave open whether the improving recipes depend
on different internal machinery. Across the mechanisms we measure, the audited
recipes show substantial reuse. The audited SFT and DPO arms largely recover
the starting model's causal head set. For conflict-SFT and DPO, subtracting a
grounding direction estimated from the starting model suppresses the gains.

\begin{table}[t]
\centering\small
\begin{tabular}{@{}lcc@{}}
\toprule
& \multicolumn{2}{c}{Overlap with base top-8} \\
Arm & 1.5B (336 heads) & 3B (576 heads) \\
\midrule
GRPO-A & 8/8 & 8/8 \\
GRPO-A$'$ & 8/8 & --- \\
SFT-E2 & 7/8 & --- \\
SFT-E0 & 7/8 & 8/8 \\
SFT-E3 & 7/8 & 8/8 \\
DPO & 7/8 & 8/8 \\
DPO seeds 1--3 & 7/8 each & --- \\
\midrule
Recall control & \textbf{0/8} & \textbf{0/8} \\
\bottomrule
\end{tabular}
\caption{\textbf{Audited arms largely recover the starting model's top causal
heads.} Top-8 causal heads are discovered independently per arm by per-head
knockout. Overlap with the starting model is 7--8/8 in the reported arms, with
hypergeometric $p\le7.1\mathrm{e}{-13}$ at 1.5B and
$\le3.5\mathrm{e}{-18}$ at 3B. A matched parametric-recall task run through
the same pipeline shares zero heads.}
\label{tab:heads}
\end{table}

\paragraph{Training largely preserves the causal head set.}
Per-head knockout over all heads, run independently per arm, recovers the base
model's top-8 head set in every swept arm at the scales tested (Table~\ref{tab:heads}, where the sweeps cover both seeded GRPO variants, all three SFT arms, and DPO,
while variants B, C, and D were not swept).
Per-seed DPO discovery gives 7/8 each, and at 7B the base-vs-DPO overlap is 7/8 of 784.
Because head-set overlap alone can be high even between unrelated tasks
\citep{merullo2024circuit,circuit_notspecific_2026}, a stronger control is
\textbf{cross-task ablation asymmetry}.
Removing the conflict heads collapses the conflict task (mean logit-difference drop $6.76$, sign-flip rate $.843$) but barely affects a matched recall task (drop $0.35$, flips $.008$).
Recall heads show the opposite pattern (recall task drop $6.79$, conflict task $1.90$).
The set also exceeds a 100-random-8-head-set permutation null on directional signal (observed drop $.638$, beyond all 100 draws, null $.141{\pm}.108$).
For DPO, seed-instability of head importance \citep{prasanna2020bert} is checked
by repeating head discovery across seeds, and identity-vs-algorithm concerns \citep{tigges2024llm} by the interventions below, which do not depend on head identity.
In short, the audited arms largely recover the same small causal head set as
the starting model, while the recall controls show that this overlap is not
generic across the two tasks.

\paragraph{The base-estimated direction remains aligned across trained arms.}
The DiM direction estimated on the base remains closely aligned with the
directions estimated from the trained models.
Its cosine to four trained arms' own directions spans $.915$--$.984$ at 1.5B, and stays between $.942$ and $.987$ at 3B, 7B, and Llama.
DPO-seed directions are equally aligned ($.950$--$.974$).
A probe at the same layer reaches AUROC $.818$ with Hewitt-Liang selectivity
$.339$ \citep{hewitt2019designing}. The layer selected by probe selectivity
also supports a causal intervention.

\begin{table}[t]
\centering\small
\setlength{\tabcolsep}{4pt}
\begin{tabular}{@{}llrrrr@{}}
\toprule
Target & Interv. & DiM & rand. & effect & $p$ \\
\midrule
base 1.5B & $+20$ lift & .569 & .460 & $+.109$ & $2.7\mathrm{e}{-4}$ \\
DPO 1.5B & $-50$ rem. & .425 & .701 & $-.276$ & $2.5\mathrm{e}{-8}$ \\
E3 1.5B & $-50$ rem. & .267 & .672 & $-.405$ & $2.2\mathrm{e}{-16}$ \\
DPO 3B & $-50$, L22 & .493 & .655 & $-.162$ & $1.9\mathrm{e}{-4}$ \\
\midrule
base 7B & $+25$ lift & .434 & .337 & $+.096$ & $3.1\mathrm{e}{-5}$ \\
DPO 7B & $-62$, L18 & .526 & .701 & $-.175$ & $3.5\mathrm{e}{-6}$ \\
DPO Llama & $-13$, L15 & .595 & .792 & $-.196$ & $2.5\mathrm{e}{-7}$ \\
DPO Llama & $-13$, L21 & .780 & .798 & $-.018$ & .375 \\
\bottomrule
\end{tabular}
\caption{\textbf{Subtracting the starting-model direction suppresses trained
grounding gains.} Injection and removal use the base-estimated DiM and are
compared with matched-norm random directions (exclusive follow-ctx rate over
frozen known sets, paired McNemar). At Llama, L15 is effective while L21 is
not, showing that the effective intervention layer can differ across model
families.}
\label{tab:intv}
\end{table}

\paragraph{The starting-model direction lifts the base model and suppresses
SFT and DPO gains.}
Adding the base direction to the base model lifts grounding ($+.109$ over
matched random). Subtracting it from DPO removes most of the measured gain
($-.276$), and subtracting it from conflict-SFT removes nearly all of it
($-.405$). DPO suppression also appears at 3B ($-.162$), 7B ($-.175$), and
Llama ($-.196$ at L15; Table~\ref{tab:intv}). Qwen-scale intervention
strengths are $75$--$190\%$ of Llama's residual norm and severely damage
generation, motivating norm calibration across families. Non-decisive
fractions rise under suppression (e.g.\ $.41$ vs.\ $.27$ for DPO-1.5B), so
suppression also reduces decisiveness, which we report alongside the effects.
The direction therefore causally affects grounding. At the selected lift doses, the closed-book pass changes
little (Appendix~\ref{app:sidefx}), supporting an effect on source selection
rather than a simple loss of parametric knowledge.

\section{Early Gains on a Stable Direction}
\label{sec:traj}

\begin{figure}[t]
\centering
\includegraphics[width=\columnwidth]{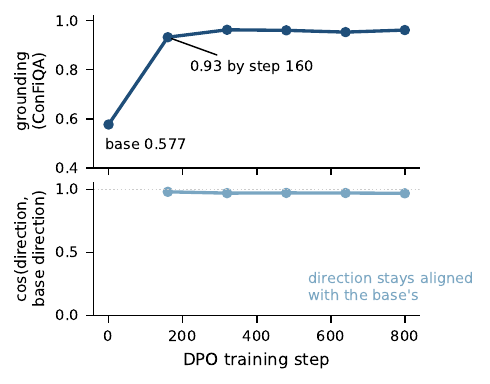}
\caption{\textbf{Grounding is largely in place by the first checkpoint, and the audited direction stays aligned (cosine $\ge.968$).} Top: ConFiQA grounding jumps from base $.577$ to $.932$ by DPO step 160 of 800 (matched 500-item evaluation). Bottom: at every checkpoint, the grounding direction's cosine to the base model's own direction stays above $.968$ (full $0$--$1$ axis).}
\label{fig:traj}
\end{figure}

To test whether the measured grounding direction changes as DPO improves
grounding, we track it throughout training. It remains closely aligned with its
starting orientation (Figure~\ref{fig:traj}).

ConFiQA grounding reaches $.932$ by step 160 of 800 (${\sim}90\%$ of the
final $.961$, base $.577$, same 500-item evaluation), while the DiM direction
at each checkpoint stays aligned with the base direction: cosine
$+.980/.970/.972/.971/.968$ at steps 160--800.

Grounding therefore rises rapidly while the measured direction remains closely
aligned with its starting orientation. This is consistent with an
amplification or recruitment account \citep{prakash2024finetuning}: most of the
behavioural gain is already present by 20\% of training without a large rotation
of the measured direction.

\section{Testing a Coverage Explanation}
\label{sec:coverage}

\begin{figure}[t]
\centering
\includegraphics[width=\columnwidth]{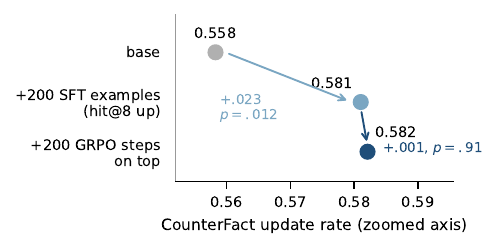}
\caption{\textbf{A supervised warm start does not rescue the tested GRPO
recipe.} The warm start raises context-answer hit@8 and grounding
($+.023$, $p{=}.012$), but the advantage-collapsed fraction remains near
$.62$. Running the identical GRPO recipe on top adds $+.001$ ($p{=}.91$).
Dot position encodes the update rate on a zoomed axis.}
\label{fig:coverage}
\end{figure}

A natural explanation for the small GRPO gains is limited exploration.
If rollouts rarely sample the context answer, GRPO may receive little useful
signal \citep{yue2025does}. We test this with rollout statistics and a
supervised warm start (Figure~\ref{fig:coverage}).

On evaluation items, context-answer coverage is already high: hit@8 $=.978$,
with only 6.0\% of items advantage-collapsed. On the training distribution, however, 61.3\% of items are
advantage-collapsed. A 200-example SFT warm start raises hit@8
from $.380$ to $.453$ and grounding by $+.023$ ($p{=}.012$), while the
advantage-collapsed fraction remains at about $.62$. Running the identical GRPO
recipe from this warm start then adds only $+.001$ ($p{=}.91$).

The warm start therefore improves one measure of context-answer coverage but
does not remove the lack of within-group reward variation. It is not sufficient
to make the tested GRPO recipe improve grounding. This experiment does not rule
out stronger coverage interventions.

\section{Steering as Causal Validation, With Limits}
\label{sec:steer}

\begin{figure}[t]
\centering
\includegraphics[width=\columnwidth]{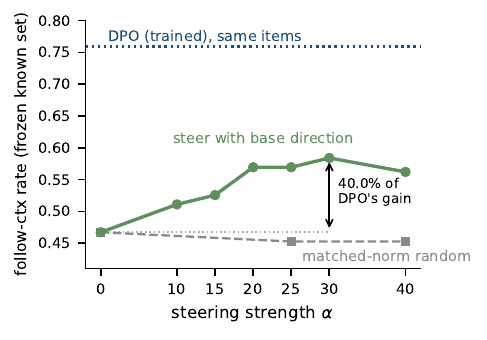}
\caption{\textbf{Steering along the base-estimated direction increases grounding. Matched random directions do not.} Steering the base model along the base-estimated direction raises follow-ctx steadily up to $\alpha{=}30$, recovering 40.0\% of the gain DPO achieves on the identical items ($35.0\%$ at the largest tested dose that passes all stated side-effect checks). Matched-norm random directions stay flat.}
\label{fig:dose}
\end{figure}

If training amplifies a pre-existing direction, steering along that direction should reproduce part of the trained behaviour.
It does, within limits (Figure~\ref{fig:dose}).
All values below are exclusive follow-ctx on the frozen base known set ($n{=}137$).
The response rises steadily with dose up to $\alpha{=}30$ ($.467$ at $\alpha{=}0$ to $.584$, matched randoms flat at $.453$), then declines mildly at $\alpha{=}40$ ($.562$).

Defining recovery as $(m_{\text{steer}}-m_{\text{base}})/(m_{\text{DPO}}-m_{\text{base}})$ on identical items, steering recovers \textbf{40.0\%} of DPO's gain with zero training.
Related work also reports partial recovery of post-training gains through
steering \citep{venhoff2026steering}, and since a ratio depends on both of its endpoints we also report the three absolute rates: $.584$, $.467$, $.759$.
Per-input reporting \citep{tan2024analysing} shows that at $\alpha{=}20$ steering helps 10.2\% of items and hurts \textbf{0.0\%} (at $\alpha{=}25$, 11.7\% and 1.5\%, matched random 1.5\% and 2.9\%).
KL at the intervened position is $.076$ at $\alpha{=}20$, below the $.1$
threshold used by \citet{arditi2024refusal}, and the MMLU and ARC checks move
within about one standard error (full dose sweep and capability numbers in Appendix~\ref{app:sidefx}).
At the 40.0\%-recovery dose ($\alpha{=}30$) the KL is $.170$, over that
threshold, so the largest tested dose passing all stated side-effect checks is
$\alpha{=}20$, which recovers \textbf{35.0\%}.
Within AxBench's steering-vs-prompting-vs-fine-tuning frame
\citep{axbench2025}, steering and context-faithful prompting
\citep{zhou2023context} are close ($.569$ vs.\ $.555$). Adding steering on top
of the prompt is not significant ($.555{\to}.606$, $p{=}.118$), and DPO remains
far stronger ($.759$ on the same items).
Steering stacked on prompting was already shown by \citet{rimsky2024caa}, we make no additivity claim, and this section validates the pre-existing-direction account rather than proposing to replace training.

\section{Metric Validation}
\label{sec:judge}

We assess our lexical-containment metric against an LLM judge (Qwen2.5-14B) on 400 items covering base and DPO outputs, two datasets, and four intervention conditions (final iteration: $\kappa{=}.507$).
Agreement is higher on training-arm than intervention outputs (final-iteration means $.73$ vs.\ $.58$), because intervention generations often mention an answer without committing to it (Appendix~\ref{app:judge}).
As a metric-artifact check \citep{schaeffer2023emergent}, the base-to-DPO contrast reproduces under judge labels, with follow-ctx rising from $.45$ to $.61$ on CounterFact and from $.46$ to $.70$ on ConFiQA.

\section{Conclusion}
\label{sec:conc}

Across nine arms (six audited mechanistically), three model families, and three
datasets, the grounding gains we measure largely depend on machinery already
present in the instruction-tuned starting model. The tested GRPO recipes yield
small grounding gains, conflict-SFT improves grounding moderately, and DPO
produces the largest gains, reaching near-ceiling grounding on its matched
distribution. For conflict-SFT and DPO, subtracting the starting-model
grounding direction suppresses the gains, while the audited models largely
recover the starting model's causal head set. \textbf{In our setting, these
results support a mechanism-reuse account of context-grounding post-training
and show that useful parts of this machinery can be identified before
training.}\label{conc:end}

\section*{Limitations}
\label{sec:lim}
\noindent\hangindent=1.1em\hangafter=1 $\bullet$~\textbf{Measurement.} Our metric is lexical containment. It agrees only moderately with an LLM judge ($\kappa{=}.507$), and 100 blind hand-labels from one author favour the judge (Appendix~\ref{app:judge}), so we rely on contrasts rather than absolute rates, and the SFT and GRPO contrasts rest on the lexical metric alone.

\noindent\hangindent=1.1em\hangafter=1 $\bullet$~\textbf{Scope.} Mechanistic audits cover the six arms in \S\ref{sec:mech} (B, C, D behavioural only), interventions are single-seed, we claim causal head sets rather than complete circuits, with the standard caveats on circuit completeness and interpretability illusions \citep{wang2023interpretability,mib2025,makelov2024subspace,circuit_variance_2026}, and experiments extend to 7B and one behaviour domain.

\noindent\hangindent=1.1em\hangafter=1 $\bullet$~\textbf{Claims.} ``Pre-existing'' means present in the instruction-tuned starting checkpoint, the arms compare complete recipes in which objective, data, and budget vary together, and stability of one direction and one head set cannot rule out new computation on axes we did not measure.

\noindent\hangindent=1.1em\hangafter=1 $\bullet$~\textbf{GRPO bound.} It covers on-policy full-model training with our reward family from a base policy with low context-answer rollout coverage in training. Synthetic-coverage \citep{si2025canoe} and frozen-backbone gate-module recipes \citep{shift2026} fall outside it.

\section*{Ethics Statement}
We study how training methods change evidence-following in open-weight models (Qwen, Llama, Phi) on public datasets (CounterFact, ConFiQA, FaithEval, HotpotQA).
No human subjects or personal data are involved.
Steering directions that raise or suppress context-following are dual-use in the narrow sense that suppression degrades faithfulness.
The effect requires white-box access, is well within what fine-tuning already enables, and our results support auditing deployed models.

\bibliography{grounding}

@inproceedings{longpre2021entity,
    title = "Entity-Based Knowledge Conflicts in Question Answering",
    author = "Longpre, Shayne  and
      Perisetla, Kartik  and
      Chen, Anthony  and
      Ramesh, Nikhil  and
      DuBois, Chris  and
      Singh, Sameer",
    editor = "Moens, Marie-Francine  and
      Huang, Xuanjing  and
      Specia, Lucia  and
      Yih, Scott Wen-tau",
    booktitle = "Proceedings of the 2021 Conference on Empirical Methods in Natural Language Processing",
    month = nov,
    year = "2021",
    address = "Online and Punta Cana, Dominican Republic",
    publisher = "Association for Computational Linguistics",
    url = "https://aclanthology.org/2021.emnlp-main.565/",
    doi = "10.18653/v1/2021.emnlp-main.565",
    pages = "7052--7063"
}

@inproceedings{
ming2025faitheval,
title={FaithEval: Can Your Language Model Stay Faithful to Context, Even If ''The Moon is Made of Marshmallows''},
author={Yifei Ming and Senthil Purushwalkam and Shrey Pandit and Zixuan Ke and Xuan-Phi Nguyen and Caiming Xiong and Shafiq Joty},
booktitle={The Thirteenth International Conference on Learning Representations},
year={2025},
url={https://openreview.net/forum?id=UeVx6L59fg}
}

@inproceedings{bi2024contextdpo,
    title = "Context-{DPO}: Aligning Language Models for Context-Faithfulness",
    author = "Bi, Baolong  and
      Huang, Shaohan  and
      Wang, Yiwei  and
      Yang, Tianchi  and
      Zhang, Zihan  and
      Huang, Haizhen  and
      Mei, Lingrui  and
      Fang, Junfeng  and
      Li, Zehao  and
      Wei, Furu  and
      Deng, Weiwei  and
      Sun, Feng  and
      Zhang, Qi  and
      Liu, Shenghua",
    editor = "Che, Wanxiang  and
      Nabende, Joyce  and
      Shutova, Ekaterina  and
      Pilehvar, Mohammad Taher",
    booktitle = "Findings of the Association for Computational Linguistics: ACL 2025",
    month = jul,
    year = "2025",
    address = "Vienna, Austria",
    publisher = "Association for Computational Linguistics",
    url = "https://aclanthology.org/2025.findings-acl.536/",
    doi = "10.18653/v1/2025.findings-acl.536",
    pages = "10280--10300",
    ISBN = "979-8-89176-256-5"
}

@inproceedings{zhou2023context,
    title = "Context-faithful Prompting for Large Language Models",
    author = "Zhou, Wenxuan  and
      Zhang, Sheng  and
      Poon, Hoifung  and
      Chen, Muhao",
    editor = "Bouamor, Houda  and
      Pino, Juan  and
      Bali, Kalika",
    booktitle = "Findings of the Association for Computational Linguistics: EMNLP 2023",
    month = dec,
    year = "2023",
    address = "Singapore",
    publisher = "Association for Computational Linguistics",
    url = "https://aclanthology.org/2023.findings-emnlp.968/",
    doi = "10.18653/v1/2023.findings-emnlp.968",
    pages = "14544--14556"
}

@inproceedings{shi2024trusting,
    title = "Trusting Your Evidence: Hallucinate Less with Context-aware Decoding",
    author = "Shi, Weijia  and
      Han, Xiaochuang  and
      Lewis, Mike  and
      Tsvetkov, Yulia  and
      Zettlemoyer, Luke  and
      Yih, Wen-tau",
    editor = "Duh, Kevin  and
      Gomez, Helena  and
      Bethard, Steven",
    booktitle = "Proceedings of the 2024 Conference of the North American Chapter of the Association for Computational Linguistics: Human Language Technologies (Volume 2: Short Papers)",
    month = jun,
    year = "2024",
    address = "Mexico City, Mexico",
    publisher = "Association for Computational Linguistics",
    url = "https://aclanthology.org/2024.naacl-short.69/",
    doi = "10.18653/v1/2024.naacl-short.69",
    pages = "783--791"
}

@inproceedings{li2022kaft,
    title = "Large Language Models with Controllable Working Memory",
    author = "Li, Daliang  and
      Rawat, Ankit Singh  and
      Zaheer, Manzil  and
      Wang, Xin  and
      Lukasik, Michal  and
      Veit, Andreas  and
      Yu, Felix  and
      Kumar, Sanjiv",
    editor = "Rogers, Anna  and
      Boyd-Graber, Jordan  and
      Okazaki, Naoaki",
    booktitle = "Findings of the Association for Computational Linguistics: ACL 2023",
    month = jul,
    year = "2023",
    address = "Toronto, Canada",
    publisher = "Association for Computational Linguistics",
    url = "https://aclanthology.org/2023.findings-acl.112/",
    doi = "10.18653/v1/2023.findings-acl.112",
    pages = "1774--1793"
}

@inproceedings{
minder2025controllable,
title={Controllable Context Sensitivity and the Knob Behind It},
author={Julian Minder and Kevin Du and Niklas Stoehr and Giovanni Monea and Chris Wendler and Robert West and Ryan Cotterell},
booktitle={The Thirteenth International Conference on Learning Representations},
year={2025},
url={https://openreview.net/forum?id=Igm9bbkzHC}
}

@inproceedings{ortu2024competition,
    title = "Competition of Mechanisms: Tracing How Language Models Handle Facts and Counterfactuals",
    author = {Ortu, Francesco  and
      Jin, Zhijing  and
      Doimo, Diego  and
      Sachan, Mrinmaya  and
      Cazzaniga, Alberto  and
      Sch{\"o}lkopf, Bernhard},
    editor = "Ku, Lun-Wei  and
      Martins, Andre  and
      Srikumar, Vivek",
    booktitle = "Proceedings of the 62nd Annual Meeting of the Association for Computational Linguistics (Volume 1: Long Papers)",
    month = aug,
    year = "2024",
    address = "Bangkok, Thailand",
    publisher = "Association for Computational Linguistics",
    url = "https://aclanthology.org/2024.acl-long.458/",
    doi = "10.18653/v1/2024.acl-long.458",
    pages = "8420--8436"
}

@inproceedings{jin2024cutting,
    title = "Cutting Off the Head Ends the Conflict: A Mechanism for Interpreting and Mitigating Knowledge Conflicts in Language Models",
    author = "Jin, Zhuoran  and
      Cao, Pengfei  and
      Yuan, Hongbang  and
      Chen, Yubo  and
      Xu, Jiexin  and
      Li, Huaijun  and
      Jiang, Xiaojian  and
      Liu, Kang  and
      Zhao, Jun",
    editor = "Ku, Lun-Wei  and
      Martins, Andre  and
      Srikumar, Vivek",
    booktitle = "Findings of the Association for Computational Linguistics: ACL 2024",
    month = aug,
    year = "2024",
    address = "Bangkok, Thailand",
    publisher = "Association for Computational Linguistics",
    url = "https://aclanthology.org/2024.findings-acl.70/",
    doi = "10.18653/v1/2024.findings-acl.70",
    pages = "1193--1215"
}

@misc{contextfocus2026,
      title={ContextFocus: Activation Steering for Contextual Faithfulness in Large Language Models}, 
      author={Nikhil Anand and Shwetha Somasundaram and Anirudh Phukan and Apoorv Saxena and Koyel Mukherjee},
      year={2026},
      eprint={2601.04131},
      archivePrefix={arXiv},
      primaryClass={cs.CL},
      url={https://arxiv.org/abs/2601.04131}, 
}

@inproceedings{
yue2025does,
title={Does Reinforcement Learning Really Incentivize Reasoning Capacity in {LLM}s Beyond the Base Model?},
author={Yang Yue and Zhiqi Chen and Rui Lu and Andrew Zhao and Zhaokai Wang and Yang Yue and Shiji Song and Gao Huang},
booktitle={The Thirty-ninth Annual Conference on Neural Information Processing Systems},
year={2025},
url={https://openreview.net/forum?id=4OsgYD7em5}
}

@misc{invisibleleash2025,
      title={The Invisible Leash: Why RLVR May or May Not Escape Its Origin}, 
      author={Fang Wu and Weihao Xuan and Ximing Lu and Mingjie Liu and Yi Dong and Zaid Harchaoui and Yejin Choi},
      year={2026},
      eprint={2507.14843},
      archivePrefix={arXiv},
      primaryClass={cs.LG},
      url={https://arxiv.org/abs/2507.14843}, 
}

@inproceedings{
shao2025spurious,
title={Spurious Rewards: Rethinking Training Signals in {RLVR}},
author={Rulin Shao and Shuyue Stella Li and Rui Xin and Scott Geng and Yiping Wang and Sewoong Oh and Simon Shaolei Du and Nathan Lambert and Sewon Min and Ranjay Krishna and Yulia Tsvetkov and Hannaneh Hajishirzi and Pang Wei Koh and Luke Zettlemoyer},
booktitle={Forty-third International Conference on Machine Learning},
year={2026},
url={https://openreview.net/forum?id=tqTNOpkP5j}
}

@inproceedings{
liu2025drgrpo,
title={Understanding R1-Zero-Like Training: A Critical Perspective},
author={Zichen Liu and Changyu Chen and Wenjun Li and Penghui Qi and Tianyu Pang and Chao Du and Wee Sun Lee and Min Lin},
booktitle={Second Conference on Language Modeling},
year={2025},
url={https://openreview.net/forum?id=5PAF7PAY2Y}
}

@inproceedings{si2025canoe,
author = {Si, Shuzheng and Zhao, Haozhe and Gao, Cheng and Bai, Yuzhuo and Wang, Zhitong and Gao, Bofei and Luo, Kangyang and Li, Wenhao and Huang, Yufei and Chen, Gang and Qi, Fanchao and Zhang, Minjia and Chang, Baobao and Sun, Maosong},
title = {Teaching large language models to maintain contextual faithfulness via synthetic tasks and reinforcement learning},
year = {2026},
isbn = {978-1-57735-906-7},
publisher = {AAAI Press},
url = {https://doi.org/10.1609/aaai.v40i39.40582},
doi = {10.1609/aaai.v40i39.40582},
booktitle = {Proceedings of the Fortieth AAAI Conference on Artificial Intelligence and Thirty-Eighth Conference on Innovative Applications of Artificial Intelligence and Sixteenth Symposium on Educational Advances in Artificial Intelligence},
articleno = {3681},
numpages = {9},
series = {AAAI'26/IAAI'26/EAAI'26}
}

@misc{cot_grpo_2025,
      title={Evaluating GRPO and DPO for Faithful Chain-of-Thought Reasoning in LLMs}, 
      author={Hadi Mohammadi and Tamas Kozak and Anastasia Giachanou},
      year={2025},
      eprint={2512.22631},
      archivePrefix={arXiv},
      primaryClass={cs.CL},
      url={https://arxiv.org/abs/2512.22631}, 
}

@misc{context_usage_2026,
      title={Emergence of Context Characteristics Sensitivity in Large Language Models}, 
      author={Nadya Yuki Wangsajaya and Haeun Yu and Isabelle Augenstein},
      year={2026},
      eprint={2606.09525},
      archivePrefix={arXiv},
      primaryClass={cs.CL},
      url={https://arxiv.org/abs/2606.09525}, 
}

@inproceedings{
weightspace_2026,
title={Weight-Space Geometry of Offline Reasoning Training},
author={Aleksandr Nikolich and Igor Kiselev and Vladimir Platonov and Karina Romanova},
booktitle={Mechanistic Interpretability Workshop at ICML 2026},
year={2026},
url={https://openreview.net/forum?id=mzgEXubB5M}
}

@misc{dim_specificity_2026,
      title={Actionable Activation Directions for Detecting and Mitigating Emergent Misalignment Across Language Model Families}, 
      author={Abdul Rafay Syed},
      year={2026},
      eprint={2606.20225},
      archivePrefix={arXiv},
      primaryClass={cs.CL},
      url={https://arxiv.org/abs/2606.20225}, 
}

@inproceedings{
circuit_notspecific_2026,
title={How Much Do Circuits Tell Us? Measuring the Consistency and Specificity of Language Model Circuits},
author={Michael Li and Nishant Subramani},
booktitle={Workshop on Scientific Understanding of Foundation Models},
year={2026},
url={https://openreview.net/forum?id=gE8q1TxmVi}
}

@misc{circuit_variance_2026,
      title={Demystifying Variance in Circuit Discovery of LLMs}, 
      author={Frank Zhengqing Wu and Francesco Tonin and Volkan Cevher},
      year={2026},
      eprint={2606.16920},
      archivePrefix={arXiv},
      primaryClass={cs.LG},
      url={https://arxiv.org/abs/2606.16920}, 
}

@misc{vision_default_2026,
      title={Vision-Default, Prior-Override: Causal Mechanisms of Perception-Knowledge Conflict in Vision-Language Models}, 
      author={Niclas Lietzow and Danielle Bitterman and Carsten Eickhoff and William Rudman and Michal Golovanevsky},
      year={2026},
      eprint={2606.28273},
      archivePrefix={arXiv},
      primaryClass={cs.CL},
      url={https://arxiv.org/abs/2606.28273}, 
}

@inproceedings{
harmfulness_refusal_2025,
title={{LLM}s Encode Harmfulness and Refusal Separately},
author={Jiachen Zhao and Jing Huang and Zhengxuan Wu and David Bau and Weiyan Shi},
booktitle={The Thirty-ninth Annual Conference on Neural Information Processing Systems},
year={2025},
url={https://openreview.net/forum?id=zLkpt30ngy}
}

@inproceedings{probing_prediction_2025,
    title = "Mechanisms vs. Outcomes: Probing for Syntax Fails to Explain Performance on Targeted Syntactic Evaluations",
    author = "Agarwal, Ananth  and
      Jian, Jasper  and
      Manning, Christopher D  and
      Murty, Shikhar",
    editor = "Christodoulopoulos, Christos  and
      Chakraborty, Tanmoy  and
      Rose, Carolyn  and
      Peng, Violet",
    booktitle = "Proceedings of the 2025 Conference on Empirical Methods in Natural Language Processing",
    month = nov,
    year = "2025",
    address = "Suzhou, China",
    publisher = "Association for Computational Linguistics",
    url = "https://aclanthology.org/2025.emnlp-main.1712/",
    doi = "10.18653/v1/2025.emnlp-main.1712",
    pages = "33737--33757",
    ISBN = "979-8-89176-332-6"
}

@inproceedings{
venhoff2026steering,
title={Base Models Know How to Reason, Thinking Models Learn When},
author={Constantin Venhoff and Iv{\'a}n Arcuschin and Philip Torr and Arthur Conmy and Neel Nanda},
booktitle={Forty-third International Conference on Machine Learning},
year={2026},
url={https://openreview.net/forum?id=2BniakOS4q}
}

@inproceedings{
prakash2024finetuning,
title={Fine-Tuning Enhances Existing Mechanisms: A Case Study on Entity Tracking},
author={Nikhil Prakash and Tamar Rott Shaham and Tal Haklay and Yonatan Belinkov and David Bau},
booktitle={The Twelfth International Conference on Learning Representations},
year={2024},
url={https://openreview.net/forum?id=8sKcAWOf2D}
}

@inproceedings{
lee2024mechanistic,
title={A Mechanistic Understanding of Alignment Algorithms: A Case Study on {DPO} and Toxicity},
author={Andrew Lee and Xiaoyan Bai and Itamar Pres and Martin Wattenberg and Jonathan K. Kummerfeld and Rada Mihalcea},
booktitle={Forty-first International Conference on Machine Learning},
year={2024},
url={https://openreview.net/forum?id=dBqHGZPGZI}
}

@inproceedings{
merullo2024circuit,
title={Circuit Component Reuse Across Tasks in Transformer Language Models},
author={Jack Merullo and Carsten Eickhoff and Ellie Pavlick},
booktitle={The Twelfth International Conference on Learning Representations},
year={2024},
url={https://openreview.net/forum?id=fpoAYV6Wsk}
}

@inproceedings{prasanna2020bert,
    title = "{W}hen {BERT} {P}lays the {L}ottery, {A}ll {T}ickets {A}re {W}inning",
    author = "Prasanna, Sai  and
      Rogers, Anna  and
      Rumshisky, Anna",
    editor = "Webber, Bonnie  and
      Cohn, Trevor  and
      He, Yulan  and
      Liu, Yang",
    booktitle = "Proceedings of the 2020 Conference on Empirical Methods in Natural Language Processing (EMNLP)",
    month = nov,
    year = "2020",
    address = "Online",
    publisher = "Association for Computational Linguistics",
    url = "https://aclanthology.org/2020.emnlp-main.259/",
    doi = "10.18653/v1/2020.emnlp-main.259",
    pages = "3208--3229"
}

@inproceedings{tigges2024llm,
 author = {Tigges, Curt and Hanna, Michael and Yu, Qinan and Biderman, Stella},
 booktitle = {Advances in Neural Information Processing Systems},
 doi = {10.52202/079017-1287},
 editor = {A. Globerson and L. Mackey and D. Belgrave and A. Fan and U. Paquet and J. Tomczak and C. Zhang},
 pages = {40699--40731},
 publisher = {Curran Associates, Inc.},
 title = {LLM Circuit Analyses Are Consistent Across Training and Scale},
 url = {https://proceedings.neurips.cc/paper_files/paper/2024/file/47c7edadfee365b394b2a3bd416048da-Paper-Conference.pdf},
 volume = {37},
 year = {2024}
}

@InProceedings{vonrutte2024language,
  title = 	 {A Language Model’s Guide Through Latent Space},
  author =       {Von R\"{u}tte, Dimitri and Anagnostidis, Sotiris and Bachmann, Gregor and Hofmann, Thomas},
  booktitle = 	 {Proceedings of the 41st International Conference on Machine Learning},
  pages = 	 {49655--49687},
  year = 	 {2024},
  editor = 	 {Salakhutdinov, Ruslan and Kolter, Zico and Heller, Katherine and Weller, Adrian and Oliver, Nuria and Scarlett, Jonathan and Berkenkamp, Felix},
  volume = 	 {235},
  series = 	 {Proceedings of Machine Learning Research},
  month = 	 {21--27 Jul},
  publisher =    {PMLR},
  url = 	 {https://proceedings.mlr.press/v235/von-rutte24a.html}
}

@article{olsson2022context,
   title={In-context Learning and Induction Heads},
   author={Olsson, Catherine and Elhage, Nelson and Nanda, Neel and Joseph, Nicholas and DasSarma, Nova and Henighan, Tom and Mann, Ben and Askell, Amanda and Bai, Yuntao and Chen, Anna and Conerly, Tom and Drain, Dawn and Ganguli, Deep and Hatfield-Dodds, Zac and Hernandez, Danny and Johnston, Scott and Jones, Andy and Kernion, Jackson and Lovitt, Liane and Ndousse, Kamal and Amodei, Dario and Brown, Tom and Clark, Jack and Kaplan, Jared and McCandlish, Sam and Olah, Chris},
   year={2022},
   journal={Transformer Circuits Thread},
   note={https://transformer-circuits.pub/2022/in-context-learning-and-induction-heads/index.html}
}

@inproceedings{hewitt2019designing,
    title = "Designing and Interpreting Probes with Control Tasks",
    author = "Hewitt, John  and
      Liang, Percy",
    editor = "Inui, Kentaro  and
      Jiang, Jing  and
      Ng, Vincent  and
      Wan, Xiaojun",
    booktitle = "Proceedings of the 2019 Conference on Empirical Methods in Natural Language Processing and the 9th International Joint Conference on Natural Language Processing (EMNLP-IJCNLP)",
    month = nov,
    year = "2019",
    address = "Hong Kong, China",
    publisher = "Association for Computational Linguistics",
    url = "https://aclanthology.org/D19-1275/",
    doi = "10.18653/v1/D19-1275",
    pages = "2733--2743"
}

@inproceedings{
schaeffer2023emergent,
title={Are Emergent Abilities of Large Language Models a Mirage?},
author={Rylan Schaeffer and Brando Miranda and Sanmi Koyejo},
booktitle={Thirty-seventh Conference on Neural Information Processing Systems},
year={2023},
url={https://openreview.net/forum?id=ITw9edRDlD}
}

@inproceedings{
chen2024preference,
title={Preference Learning Algorithms Do Not Learn Preference Rankings},
author={Angelica Chen and Sadhika Malladi and Lily H Zhang and Xinyi Chen and Qiuyi Zhang and Rajesh Ranganath and Kyunghyun Cho},
booktitle={The Thirty-eighth Annual Conference on Neural Information Processing Systems},
year={2024},
url={https://openreview.net/forum?id=YkJ5BuEXdD}
}

@misc{mcgrath2023hydra,
      title={The Hydra Effect: Emergent Self-repair in Language Model Computations}, 
      author={Thomas McGrath and Matthew Rahtz and Janos Kramar and Vladimir Mikulik and Shane Legg},
      year={2023},
      eprint={2307.15771},
      archivePrefix={arXiv},
      primaryClass={cs.LG},
      url={https://arxiv.org/abs/2307.15771}, 
}

@inproceedings{
rushing2024explorations,
title={Explorations of Self-Repair in Language Models},
author={Cody Rushing and Neel Nanda},
booktitle={Forty-first International Conference on Machine Learning},
year={2024},
url={https://openreview.net/forum?id=5ZwEifshyo}
}

@inproceedings{
wollschlager2025geometry,
title={The Geometry of Refusal in Large Language Models: Concept Cones and Representational Independence},
author={Tom Wollschl{\"a}ger and Jannes Elstner and Simon Geisler and Vincent Cohen-Addad and Stephan G{\"u}nnemann and Johannes Gasteiger},
booktitle={Forty-second International Conference on Machine Learning},
year={2025},
url={https://openreview.net/forum?id=80IwJqlXs8}
}

@inproceedings{
miller2024transformer,
title={Transformer Circuit Evaluation Metrics Are Not Robust},
author={Joseph Miller and Bilal Chughtai and William Saunders},
booktitle={First Conference on Language Modeling},
year={2024},
url={https://openreview.net/forum?id=zSf8PJyQb2}
}

@inproceedings{
wang2023interpretability,
title={Interpretability in the Wild: a Circuit for Indirect Object Identification in {GPT}-2 Small},
author={Kevin Ro Wang and Alexandre Variengien and Arthur Conmy and Buck Shlegeris and Jacob Steinhardt},
booktitle={The Eleventh International Conference on Learning Representations },
year={2023},
url={https://openreview.net/forum?id=NpsVSN6o4ul}
}

@inproceedings{
makelov2024subspace,
title={Is This the Subspace You Are Looking for? An Interpretability Illusion for Subspace Activation Patching},
author={Aleksandar Makelov and Georg Lange and Atticus Geiger and Neel Nanda},
booktitle={The Twelfth International Conference on Learning Representations},
year={2024},
url={https://openreview.net/forum?id=Ebt7JgMHv1}
}

@inproceedings{
arditi2024refusal,
title={Refusal in Language Models Is Mediated by a Single Direction},
author={Andy Arditi and Oscar Balcells Obeso and Aaquib Syed and Daniel Paleka and Nina Rimsky and Wes Gurnee and Neel Nanda},
booktitle={The Thirty-eighth Annual Conference on Neural Information Processing Systems},
year={2024},
url={https://openreview.net/forum?id=pH3XAQME6c}
}

@inproceedings{rimsky2024caa,
    title = "Steering Llama 2 via Contrastive Activation Addition",
    author = "Rimsky, Nina  and
      Gabrieli, Nick  and
      Schulz, Julian  and
      Tong, Meg  and
      Hubinger, Evan  and
      Turner, Alexander",
    editor = "Ku, Lun-Wei  and
      Martins, Andre  and
      Srikumar, Vivek",
    booktitle = "Proceedings of the 62nd Annual Meeting of the Association for Computational Linguistics (Volume 1: Long Papers)",
    month = aug,
    year = "2024",
    address = "Bangkok, Thailand",
    publisher = "Association for Computational Linguistics",
    url = "https://aclanthology.org/2024.acl-long.828/",
    doi = "10.18653/v1/2024.acl-long.828",
    pages = "15504--15522"
}

@misc{turner2023actadd,
      title={Steering Language Models With Activation Engineering}, 
      author={Alexander Matt Turner and Lisa Thiergart and Gavin Leech and David Udell and Juan J. Vazquez and Ulisse Mini and Monte MacDiarmid},
      year={2024},
      eprint={2308.10248},
      archivePrefix={arXiv},
      primaryClass={cs.CL},
      url={https://arxiv.org/abs/2308.10248}, 
}

@misc{zou2023repe,
      title={Representation Engineering: A Top-Down Approach to AI Transparency}, 
      author={Andy Zou and Long Phan and Sarah Chen and James Campbell and Phillip Guo and Richard Ren and Alexander Pan and Xuwang Yin and Mantas Mazeika and Ann-Kathrin Dombrowski and Shashwat Goel and Nathaniel Li and Michael J. Byun and Zifan Wang and Alex Mallen and Steven Basart and Sanmi Koyejo and Dawn Song and Matt Fredrikson and J. Zico Kolter and Dan Hendrycks},
      year={2025},
      eprint={2310.01405},
      archivePrefix={arXiv},
      primaryClass={cs.LG},
      url={https://arxiv.org/abs/2310.01405}, 
}

@inproceedings{
tan2024analysing,
title={Analysing the Generalisation and Reliability of Steering Vectors},
author={Daniel Chee Hian Tan and David Chanin and Aengus Lynch and Brooks Paige and Dimitrios Kanoulas and Adri{\`a} Garriga-Alonso and Robert Kirk},
booktitle={The Thirty-eighth Annual Conference on Neural Information Processing Systems},
year={2024},
url={https://openreview.net/forum?id=v8X70gTodR}
}

@misc{pres2024towards,
      title={Towards Reliable Evaluation of Behavior Steering Interventions in LLMs}, 
      author={Itamar Pres and Laura Ruis and Ekdeep Singh Lubana and David Krueger},
      year={2024},
      eprint={2410.17245},
      archivePrefix={arXiv},
      primaryClass={cs.AI},
      url={https://arxiv.org/abs/2410.17245}, 
}

@inproceedings{yang2018hotpotqa,
    title = "{H}otpot{QA}: A Dataset for Diverse, Explainable Multi-hop Question Answering",
    author = "Yang, Zhilin  and
      Qi, Peng  and
      Zhang, Saizheng  and
      Bengio, Yoshua  and
      Cohen, William  and
      Salakhutdinov, Ruslan  and
      Manning, Christopher D.",
    editor = "Riloff, Ellen  and
      Chiang, David  and
      Hockenmaier, Julia  and
      Tsujii, Jun{'}ichi",
    booktitle = "Proceedings of the 2018 Conference on Empirical Methods in Natural Language Processing",
    month = oct # "-" # nov,
    year = "2018",
    address = "Brussels, Belgium",
    publisher = "Association for Computational Linguistics",
    url = "https://aclanthology.org/D18-1259/",
    doi = "10.18653/v1/D18-1259",
    pages = "2369--2380"
}

@inproceedings{
meng2022locating,
title={Locating and Editing Factual Associations in {GPT}},
author={Kevin Meng and David Bau and Alex J Andonian and Yonatan Belinkov},
booktitle={Advances in Neural Information Processing Systems},
editor={Alice H. Oh and Alekh Agarwal and Danielle Belgrave and Kyunghyun Cho},
year={2022},
url={https://openreview.net/forum?id=-h6WAS6eE4}
}

@misc{persona_subspace_2026,
      title={Emergent Misalignment Recruits a Pre-existing Persona Subspace}, 
      author={Mohammed Suhail B Nadaf},
      year={2026},
      eprint={2607.21356},
      archivePrefix={arXiv},
      primaryClass={cs.LG},
      url={https://arxiv.org/abs/2607.21356}, 
}

@inproceedings{shi2026features,
    title = "Why Does Reinforcement Learning Generalize? A Feature-Level Mechanistic Study of Post-Training in Large Language Models",
    author = "Shi, Dan  and
      Han, Zhuowen  and
      Ostermann, Simon  and
      Jin, Renren  and
      van Genabith, Josef  and
      Xiong, Deyi",
    editor = "Liakata, Maria  and
      Moreira, Viviane P.  and
      Zhang, Jiajun  and
      Jurgens, David",
    booktitle = "Proceedings of the 64th Annual Meeting of the {A}ssociation for {C}omputational {L}inguistics (Volume 1: Long Papers)",
    month = jul,
    year = "2026",
    address = "San Diego, California, United States",
    publisher = "Association for Computational Linguistics",
    url = "https://aclanthology.org/2026.acl-long.1808/",
    doi = "10.18653/v1/2026.acl-long.1808",
    pages = "38979--39000",
    ISBN = "979-8-89176-390-6"
}

@misc{ward2025reasoning,
      title={Reasoning-Finetuning Repurposes Latent Representations in Base Models}, 
      author={Jake Ward and Chuqiao Lin and Constantin Venhoff and Neel Nanda},
      year={2025},
      eprint={2507.12638},
      archivePrefix={arXiv},
      primaryClass={cs.LG},
      url={https://arxiv.org/abs/2507.12638}, 
}

@inproceedings{
axbench2025,
title={AxBench: Steering {LLM}s? Even Simple Baselines Outperform Sparse Autoencoders},
author={Zhengxuan Wu and Aryaman Arora and Atticus Geiger and Zheng Wang and Jing Huang and Dan Jurafsky and Christopher D Manning and Christopher Potts},
booktitle={Forty-second International Conference on Machine Learning},
year={2025},
url={https://openreview.net/forum?id=K2CckZjNy0}
}

@inproceedings{
mib2025,
title={{MIB}: A Mechanistic Interpretability Benchmark},
author={Aaron Mueller and Atticus Geiger and Sarah Wiegreffe and Dana Arad and Iv{\'a}n Arcuschin and Adam Belfki and Yik Siu Chan and Jaden Fried Fiotto-Kaufman and Tal Haklay and Michael Hanna and Jing Huang and Rohan Gupta and Yaniv Nikankin and Hadas Orgad and Nikhil Prakash and Anja Reusch and Aruna Sankaranarayanan and Shun Shao and Alessandro Stolfo and Martin Tutek and Amir Zur and David Bau and Yonatan Belinkov},
booktitle={Forty-second International Conference on Machine Learning},
year={2025},
url={https://openreview.net/forum?id=sSrOwve6vb}
}

@misc{shift2026,
      title={SHIFT: Gate-Modulated Activation Steering for Knowledge Conflict Mitigation in Retrieval-Augmented Generation}, 
      author={Ruochang Li and Pengcheng Huang and Zhenghao Liu and Yukun Yan and Huiyuan Xie and Yu Gu and Ge Yu and Maosong Sun},
      year={2026},
      eprint={2606.27786},
      archivePrefix={arXiv},
      primaryClass={cs.CL},
      url={https://arxiv.org/abs/2606.27786}, 
}

@inproceedings{
chen2026longrlvr,
title={Long{RLVR}: Long-Context Reinforcement Learning Requires Verifiable Context Rewards},
author={Guanzheng Chen and Michael Qizhe Shieh and Lidong Bing},
booktitle={The Fourteenth International Conference on Learning Representations},
year={2026},
url={https://openreview.net/forum?id=omVhYvyTPJ}
}

@misc{tamo2026evidencerl,
      title={EvidenceRL: Reinforcing Evidence Consistency for Trustworthy Language Models}, 
      author={J. Ben Tamo and Yuxing Lu and Benoit L. Marteau and Micky C. Nnamdi and May D. Wang},
      year={2026},
      eprint={2603.19532},
      archivePrefix={arXiv},
      primaryClass={cs.CL},
      url={https://arxiv.org/abs/2603.19532}, 
}

@inproceedings{bigoulaeva2026patches,
    title = "Patches of Nonlinearity: Instruction Vectors in Large Language Models",
    author = "Bigoulaeva, Irina  and
      Rohweder, Jonas  and
      Dutta, Subhabrata  and
      Gurevych, Iryna",
    editor = "Liakata, Maria  and
      Moreira, Viviane P.  and
      Zhang, Jiajun  and
      Jurgens, David",
    booktitle = "Proceedings of the 64th Annual Meeting of the {A}ssociation for {C}omputational {L}inguistics (Volume 1: Long Papers)",
    month = jul,
    year = "2026",
    address = "San Diego, California, United States",
    publisher = "Association for Computational Linguistics",
    url = "https://aclanthology.org/2026.acl-long.559/",
    doi = "10.18653/v1/2026.acl-long.559",
    pages = "12209--12262",
    ISBN = "979-8-89176-390-6"
}

@misc{gupta2026worldmodels,
      title={Better World Models Can Lead to Better Post-Training Performance}, 
      author={Prakhar Gupta and Henry Conklin and Sarah-Jane Leslie and Andrew Lee},
      year={2025},
      eprint={2512.03400},
      archivePrefix={arXiv},
      primaryClass={cs.LG},
      url={https://arxiv.org/abs/2512.03400}, 
}

@misc{gupta2026biasdirections,
      title={How Does Alignment Tuning Shape Representations of Sycophancy and Related Cue-Induced Biases in LLMs?}, 
      author={Prakhar Gupta and Terry Jingchen Zhang and Florent Draye and Bernhard Schölkopf and Zhijing Jin},
      year={2026},
      eprint={2607.18114},
      archivePrefix={arXiv},
      primaryClass={cs.CL},
      url={https://arxiv.org/abs/2607.18114}, 
}

@misc{shao2024deepseekmath,
      title={DeepSeekMath: Pushing the Limits of Mathematical Reasoning in Open Language Models}, 
      author={Zhihong Shao and Peiyi Wang and Qihao Zhu and Runxin Xu and Junxiao Song and Xiao Bi and Haowei Zhang and Mingchuan Zhang and Y. K. Li and Y. Wu and Daya Guo},
      year={2024},
      eprint={2402.03300},
      archivePrefix={arXiv},
      primaryClass={cs.CL},
      url={https://arxiv.org/abs/2402.03300}, 
}

@inproceedings{
rafailov2023direct,
title={Direct Preference Optimization: Your Language Model is Secretly a Reward Model},
author={Rafael Rafailov and Archit Sharma and Eric Mitchell and Christopher D Manning and Stefano Ermon and Chelsea Finn},
booktitle={Thirty-seventh Conference on Neural Information Processing Systems},
year={2023},
url={https://openreview.net/forum?id=HPuSIXJaa9}
}

@inproceedings{
liu2025prorl,
title={Pro{RL}: Prolonged Reinforcement Learning Expands Reasoning Boundaries in Large Language Models},
author={Mingjie Liu and Shizhe Diao and Ximing Lu and Jian Hu and Xin Dong and Yejin Choi and Jan Kautz and Yi Dong},
booktitle={The Thirty-ninth Annual Conference on Neural Information Processing Systems},
year={2025},
url={https://openreview.net/forum?id=YPsJha5HXQ}
}

\appendix

\section{The Intervention-Conditioned Denominator}
\label{app:metric}
Under a steering hook, a two-pass conflict protocol has two hidden dependencies on the intervention: (i) if the closed-book pass runs under the hook, the known set itself becomes $\alpha$-dependent, and (ii) even with the known gate frozen, update rate $=$ fc/(fc$+$fm) conditions on the item being decisive, and the non-decisive fraction is intervention-dependent (e.g.\ $.41$ under DPO-suppression vs.\ $.27$ under its matched random).
Either dependency silently shifts the evaluation set toward items the intervention finds easy.
The remedy is to freeze the denominator before the intervention is applied.
Report the exclusive follow-ctx rate over the frozen no-intervention known set and the non-decisive fraction separately.
In our own audit, correcting (ii) changed a suppression effect estimate from $-.040$ to $-.162$ and removed a false bump from the dose-response curve.
We recommend the frozen-denominator report as standard practice for steered evaluations.

\section{Judge Validation Details}
\label{app:judge}

\subsection{Iterations and Disagreement Types}
Judge-prompt iterations: naive ($\kappa{=}.627$), commitment-semantics ($.517$), alias handling ($.541$), and untruncated generations ($.507$, 4-way agreement $.652$, 4 parse failures of 400).
Disagreements fall into three groups: (a) lexical BOTH from incidental later mentions of the second answer (judge typically correct), (b) judge NEITHER on answer-plus-rambling generations (judge too strict), and (c) under suppression, generations that name the context answer without committing to it (here surface form and answer commitment genuinely come apart).
Per-source agreement (final iteration): CounterFact arms $.76$--$.78$, ConFiQA base $.80$, ConFiQA DPO $.58$, interventions $.44$--$.66$.

\subsection{Author Labelling}
One author labelled 100 outputs from the validation pool (CTX, MEM, BOTH, or NEITHER), taking 80 training-arm and 20 intervention outputs, with metric-judge disagreements oversampled by design (60 of 100).
Labelling was blind to both automated labels.
The sheet showed the question, both candidate answers, and the generation, never the metric or judge label.
The result favours the judge.
Agreement with the human label is $.78$ ($\kappa{=}.602$) for the judge vs.\ $.53$ ($\kappa{=}.346$) for the lexical metric, with the gap widest on intervention outputs (judge $.95$, $\kappa{=}.897$, metric $.75$, $\kappa{=}.550$, $n{=}20$).
On the 60 contested items the human sided with the judge 39 times, the metric 14, and neither 7, and five of the seven neither cases are mode-(b) generations that state one answer and then drift into text mentioning the other, which the human scores by the first answer given.
Where metric and judge agreed, the human confirmed the shared label on 39 of 40.
The sample deliberately over-represents disagreements, so these rates describe the contested region rather than estimate overall agreement, and the labels come from a single author.

\section{Training and Evaluation Details}
\label{app:details}
This section lists the settings needed to reproduce every arm and every test.

\subsection{Training Configurations}
GRPO: TRL implementation, 200 steps, lr $3{\cdot}10^{-6}$, KL $\beta{=}.02$, 8 rollouts per prompt, temperature 1.0, max prompt length 1536 (HotpotQA distractor median ${\sim}1400$ tokens), reward weights [answer F1 1.0, citation set-F1 0.5, context-utility 0.3, format 0.1] where applicable.
The context-utility reward is implemented as a per-token mean log-probability ratio over the completion under a frozen base-model scorer, clipped for stability.
SFT: completion-only loss, 2 epochs.
E3 mixes 78\% conflict with 22\% standard QA, and E2 follows the KAFT recipe (conflict, standard, irrelevant-context).
DPO: $\beta{=}.5$, lr $5{\cdot}10^{-6}$, 3 epochs, 4{,}500 ConFiQA preference pairs.
\subsection{Evaluation Protocol}
CounterFact: 4{,}000 items scanned.
The base model answers 1{,}089 correctly closed-book at 1.5B (1{,}257 at 3B, 1{,}252 at 7B, 1{,}394 at Llama), and the known file is frozen and shared by all arms.
Interventions run on 500 ConFiQA items with frozen known sets base-1.5B $n{=}137$, DPO-1.5B $134$, base-3B $154$, DPO-3B $142$, E3 $131$, base-7B $166$, DPO-7B $154$, and DPO-Llama $168$.
All significance tests are exact two-sided McNemar on items paired by index.
Equivalence tests are two one-sided $t$ tests over 4 seeds using $.044$, the
observed conflict-SFT gain over its matched control, as the equivalence margin.
Ablation recipe, stated per \citet{miller2024transformer}: node-level, zero and mean ablation, last position, ablate-clean direction, circuit rather than complement, logit-difference metric, aggregated after ablation.
\subsection{Code and Data Release}
Code, per-item dumps, the estimated direction vectors themselves, a one-command reproduction of every figure from the released per-item records, and a verification harness (154 automated checks that match every reported number against raw per-item records) will be released on publication.

\section{Exact Prompt Formats}
\label{app:prompts}
All templates are exactly as in the released evaluation code (placeholder names ours).
CounterFact runs as \textbf{raw completion} (no chat template).
The closed-book known-set pass completes the bare relation stem (max 8 new tokens), and the conflict pass completes \texttt{"\{stem\} \{counterfactual\}.\ \{stem\}"} (max 24 new tokens).
ConFiQA and FaithEval use the model's \textbf{chat template}.
ConFiQA with context: \texttt{"\{modified\_context\}\textbackslash n\textbackslash nQuestion: \{q\}\textbackslash nAnswer with a short phrase only."} Closed book drops the context line.
The opinion-and-instruction prompting baseline follows \citet{zhou2023context}: \texttt{"Bob said, "\{modified\_context\}".\ \{q\} in Bob's opinion?\textbackslash nAnswer with a short phrase only."} (a trailing question mark is stripped from \texttt{\{q\}}).
FaithEval with context: \texttt{"Read the passage and answer the multiple-choice question.\textbackslash n\textbackslash nPassage: \{context\}\textbackslash n\textbackslash nQuestion: \{q\}\textbackslash n\{choices\}\textbackslash n\textbackslash nAnswer with a single letter (A, B, C, or D)."} The no-context pass drops the instruction line and the passage.
The judge prompt instructs grading of the answer the response commits to (``Judge ONLY the answer the response asserts as its answer \dots\ If the response opens with an answer, that IS its answer even if it rambles after''), with one sentence of reasoning then \texttt{FINAL: CTX | MEM | BOTH | NEITHER}.

\section{Consolidated Intervention Side-Effects}
\label{app:sidefx}
A steering intervention can change behaviour beyond its target metric.
We collect every side-effect measurement here (base-1.5B, DiM at L21 unless noted):
KL at the intervened position rises with dose ($.019/.076/.117/.170/.303$ at $\alpha{=}10/20/25/30/40$, matched random $.011/.041/.065/.095/.181$), crossing the $.1$ threshold used by \citet{arditi2024refusal} between $\alpha{=}20$ and $25$.
MMLU $.423{\to}.400$ and ARC $.754{\to}.752$ at $\alpha{=}20$ ($n{=}1000$ each, ${\approx}1$ SE of the difference, n.s., matched random comparable).
The closed-book pass changes little under lift at $\alpha{\le}25$
($.312$--$.324$ vs.\ $.316$ on identical items, dipping to $.298/.282$ at
$\alpha{=}30/40$), so the deployed dose does not measurably change
closed-book recall on these items.
Per-input effects: helps $10.2\%$, hurts $0.0\%$ at $\alpha{=}20$.
HotpotQA F1 under the trained arms is reported in \S\ref{sec:behav}.
Suppression costs decisiveness (non-decisive fraction $.41$ vs.\ $.27$ for DPO$-$50), reported alongside every suppression row.

\section{Negative Results and Additional Controls}
\label{sec:dfa}
This appendix collects results that do not change the main claims but bound them or warn against tempting shortcuts.

\subsection{A Two-Axis Causal Subspace}
A second causal axis exists and stays largely independent under intervention.
A second, differently-estimated axis (context-presence DiM, following \citealp{contextfocus2026}) shares about 2\% of variance with the behavioural axis (cosine $+.146$ at 1.5B, small but reliably non-orthogonal, $5.7\sigma$ from chance at $d{=}1536$).
Geometric near-orthogonality is weak evidence \citep{wollschlager2025geometry}, so we test independence by intervention, injecting one axis while continuously projecting the other out of the residual stream.
The injected axis keeps its effect in 3 of 4 cells (1.5B suppression $.754{\to}.470$, 3B lift $.234{\to}.344$, 3B suppression $.782{\to}.556$, all $p{\le}1.5\mathrm{e}{-5}$).
The 1.5B lift cell is not significant ($.504{\to}.547$, $p{=}.070$), plausibly because projection alone already lifts base from $.467$ to $.504$, leaving little room to improve.
Multi-axis structure is known for refusal \citep{wollschlager2025geometry,harmfulness_refusal_2025}.
Here we test the corresponding structure for grounding by intervention.

\subsection{Single-Layer Attribution}
Single-layer attribution suggests a stage structure that its own controls reject.
Direction-attribution at the readout layer (DFA, \citealp{arditi2024refusal}) selects late-layer ``writer'' components that do not overlap the causal heads (0/8 at four models), which at first appears to be the routing-vs-writing decomposition reported in VLMs \citep{vision_default_2026}.
Against a 100-random-set permutation null the writers' directional-signal drop ($.174$) is within the null range ($p{=}.36$, null $.141{\pm}.108$), and the ``writers'' track the chosen readout layer (L19--21 when read at L21, L10--15 at L15).
Mean ablation reproduces this, ruling out a zero-ablation artifact \citep{rushing2024explorations,mcgrath2023hydra}.
We publish the control.
Single-readout attribution should be validated against a permutation null over matched random component sets, whose spread (sd $.108$ on mean $.141$) makes one random comparison insufficient.
The causally-necessary heads pass the identical null ($.638$, beyond all 100 draws).

\subsection{Item-Level Prediction}
The base direction is a weaker item-level predictor than likelihood.
Predicting which known items DPO flips, the base model's log-prob margin outperforms projection onto the direction ($.718$ vs.\ $.636$ AUROC among items the base does not already follow), and adding the projection to a likelihood-only predictor changes AUROC negligibly, which is the pattern preference optimization should produce if it grows margins on items nearest the boundary \citep{chen2024preference}.
We report this boundary condition on representation-based prediction \citep{vonrutte2024language,probing_prediction_2025}.

\subsection{Weight-Difference Screening}
Weight-difference magnitude fails as a screen for behavioural change.
Across all 10 arms with weight-delta data, normalized weight movement correlates positively but not significantly with behavioural effect (Pearson $+.544$, Spearman $+.600$, $n{=}10$), and a weight-diff screen ranks both SFT controls above both DPO arms.
Norm-based screening depends on objective and optimizer choices \citep{weightspace_2026}.
These results suggest that audits should include behavioural or
interventional evidence rather than relying on weight-difference magnitude
alone.

\section{Full FaithEval Results}
\label{app:faitheval}
Update rate on FaithEval counterfactual MCQA (fixed letter parser, single seed).
Llama uses the 16-token regeneration (\S\ref{sec:behav}).
The 4-token reparse gives the same qualitative pattern (base lowest, conflict-trained arms highest).
We make no claim about the E3-vs-DPO ordering on this dataset, because it is not stable across parser regimes and known-set definitions (16-token, known-in-both, $p{=}.90$, a tie).
Paired contrasts (follow-ctx, known-in-both McNemar): GRPO-A vs.\ base $-.002$ ($p{=}.77$) at 1.5B and $-.001$ ($p{=}1.0$) at 3B, E3 vs.\ E0 $+.024$ ($p{=}7\mathrm{e}{-4}$) and $+.026$ ($p{=}4\mathrm{e}{-4}$).
The near-zero GRPO-A effect and the supervised effect both replicate on a
second dataset in a different format.

\begin{table}[h]
\centering\small
\begin{tabular}{@{}lccc@{}}
\toprule
Arm & 1.5B & 3B & Llama-3.2-3B \\
\midrule
base & .684 & .622 & .728 \\
GRPO-A & .685 & .620 & --- \\
GRPO-A$'$ & .689 & --- & --- \\
SFT-E0 & .674 & .623 & .757 \\
SFT-E2 & .696 & --- & --- \\
SFT-E3 & .705 & .647 & .784 \\
DPO & .716 & .673 & .779 \\
\bottomrule
\end{tabular}
\caption{\textbf{The dissociation replicates on a second dataset in a different format.} FaithEval update rate. Missing cells: arm not trained at that backbone (A$'$ and E2 beyond 1.5B) or not evaluated on FaithEval (GRPO-A at Llama, whose CounterFact null is reported in \S\ref{sec:behav}).}
\label{tab:faitheval}
\end{table}

\section{Quantitative Summary of the Dissociation and the Audit}
\label{app:quant}
\begin{figure*}[t]
\centering
\includegraphics[width=\textwidth]{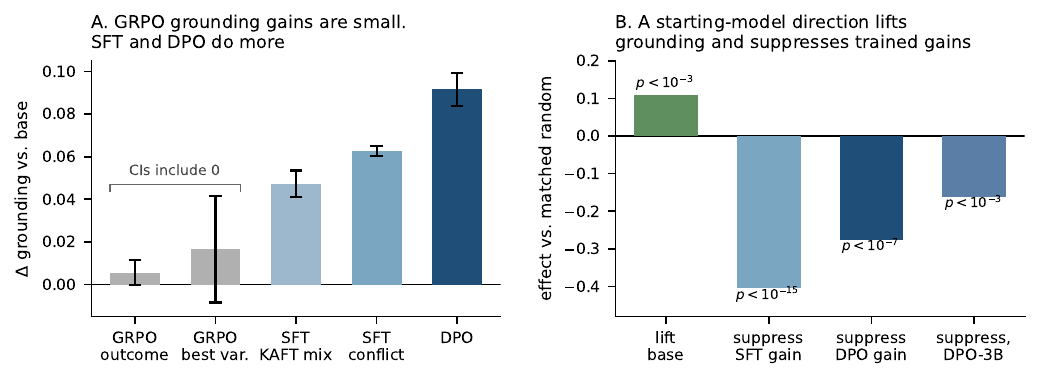}
\caption{\textbf{The dissociation and the audit, quantified.}
(A) GRPO grounding gains are small. Neither seeded GRPO variant shows a robust
gain under replication (CIs include zero, with effects equivalence-bounded
within the conflict-SFT benchmark), and the single-seed variants remain below
$+.02$. Conflict-supervised SFT improves grounding moderately, while DPO
improves it most (CounterFact, 4-seed mean $\Delta$, $t$-CIs).
(B) A difference-in-means direction estimated on the starting model lifts base
grounding and suppresses conflict-SFT and DPO gains relative to matched-norm
random directions (exclusive follow-ctx, frozen known sets).}
\label{fig:quant}
\end{figure*}

Figure~\ref{fig:quant} shows the behavioural dissociation and the direction audit side by side.

\section{Causal Head Identities}
\label{app:heads}
Base-model top-8 causal head sets (layer.head), discovered by per-head knockout and reproduced 7--8/8 by every swept arm (Table~\ref{tab:heads}):
Qwen2.5-1.5B: L8.H3 (dominant), L13.H4, L6.H6, L2.H3, L15.H7, L27.H2, L15.H6, L14.H0.
Qwen2.5-3B: L5.H9, L5.H12, L20.H8, L19.H11, L24.H1, L24.H3, L24.H10, L23.H10 (identical top-8 in all five 3B arms).
The set is also not reducible to induction heads \citep{olsson2022context}.
One of the eight heads overlaps the top-10 induction set, and the two highest-attribution heads rank 188th and 333rd of 336 by induction score.
Files \texttt{causal/causal\_*.json} in the release carry the top-30 per-head attributions from every arm's full sweep.

\end{document}